\documentclass{article}

\usepackage[nonatbib, preprint]{neurips_2025}

\usepackage[utf8]{inputenc} 
\usepackage[T1]{fontenc}    
\usepackage{hyperref}       
\usepackage{url}            
\usepackage{booktabs}       
\usepackage{amsfonts}       
\usepackage{nicefrac}       
\usepackage{microtype}      
\usepackage{natbib}

\usepackage{amsmath}
\usepackage{graphicx}

\usepackage{enumitem}

\PassOptionsToPackage{table,dvipsnames}{xcolor}\usepackage[most]{tcolorbox}
\usepackage{float}

\hypersetup{hidelinks, colorlinks=true, linkcolor=RoyalPurple, citecolor=RoyalBlue}

\title{MM-PRM: Enhancing Multimodal Mathematical Reasoning with Scalable Step-Level Supervision}

\author{
\begin{minipage}{\textwidth}
    \vspace{0.5em}
    \centering
    Lingxiao Du$^{1}$\thanks{Equal contribution} \quad Fanqing Meng$^{1,3*}$ \quad Zongkai Liu$^{2*}$ \quad Zhixiang Zhou$^{2*}$ \\[2pt]
    Ping Luo$^{4}$ \quad Qiaosheng Zhang$^{1,2\dagger}$ \quad Wenqi Shao$^{1,2}$\thanks{Corresponding Authors: shaowenqi@pjlab.org.cn; zhangqiaosheng@pjlab.org.cn} \\[6pt]
    $^{1}$Shanghai AI Laboratory \quad $^{2}$Shanghai Innovation Institute \\ 
    $^{3}$Shanghai Jiao Tong University \quad $^{4}$The University of Hong Kong
    \vspace{-1em}
\end{minipage}
}

\begin{document}

\maketitle

\begin{abstract}
While Multimodal Large Language Models (MLLMs) have achieved impressive progress in vision-language understanding, they still struggle with complex multi-step reasoning, often producing logically inconsistent or partially correct solutions. A key limitation lies in the lack of fine-grained supervision over intermediate reasoning steps. To address this, we propose MM-PRM, a process reward model trained within a fully automated, scalable framework. We first build MM-Policy, a strong multimodal model trained on diverse mathematical reasoning data. Then, we construct MM-K12, a curated dataset of 10,000 multimodal math problems with verifiable answers, which serves as seed data. Leveraging a Monte Carlo Tree Search (MCTS)-based pipeline, we generate over 700k step-level annotations without human labeling. The resulting PRM is used to score candidate reasoning paths in the Best-of-N inference setup and achieves significant improvements across both in-domain (MM-K12 test set) and out-of-domain (OlympiadBench, MathVista, etc.) benchmarks. Further analysis confirms the effectiveness of soft labels, smaller learning rates, and path diversity in optimizing PRM performance. MM-PRM demonstrates that process supervision is a powerful tool for enhancing the logical robustness of multimodal reasoning systems. We release all our codes and data at \url{https://github.com/ModalMinds/MM-PRM}.
\end{abstract}

\section{Introduction}\label{sec: introduction}

The rapid advancement of Large Language Models (LLMs)~\cite{achiam2023gpt, bai2023qwen, team2023gemini, touvron2023llama, touvron2023llama2, grattafiori2024llama, 2023internlm, cai2024internlm2, guo2025deepseek} has significantly improved performance on a wide range of natural language processing tasks, including general reasoning and mathematical problem-solving. In parallel, the development of Multimodal Large Language Models (MLLMs)~\cite{liu2023visual, yao2024minicpm, chen2024expanding, wang2024mpo, gao2024mini, chen2024far, chen2024internvl, Qwen-VL, team2025kimi} has unlocked new capabilities in vision-language understanding, showing promising results in areas such as image captioning and visual question answering (VQA). However, despite their impressive capabilities in perception and basic reasoning, MLLMs still struggle with complex multi-step reasoning tasks, particularly in mathematics. These shortcomings often manifest as broken logical chains, inaccurate intermediate steps, or cases that produce incorrect intermediate steps while still occasionally arriving at the correct final answer---a phenomenon that introduces high false-positive rates and undermines interpretability.

To address this issue, reward modeling~\cite{cobbe2021training, yu2023ovm, uesato2022solving, zang2025internlm} has emerged as a promising paradigm. Reward models (RMs) play a central role in reinforcement learning from human feedback (RLHF)~\cite{ouyang2022training, schulman2017proximal, shao2024deepseekmath, lai2024step, wang2024enhancing, pang2024iterative}, and can also be used at inference time to select among multiple candidate responses using Test-Time Scaling (TTS)~\cite{dong2024rlhf, snell2024scaling, wang2023math, zhang2025lessons, feng2023alphazero, kang2024mindstar, ma2023let, tian2024toward, zhang2024rest} strategies such as Best-of-N (BoN). 
In general, reward models for reasoning tasks can be broadly categorized into two types: Outcome Reward Models (ORMs) and Process Reward Models (PRMs). ORMs~\cite{cobbe2021training, yu2023ovm} provide scalar feedback only on the final answer, overlooking the quality of the intermediate reasoning steps. This limits their ability to guide the model toward robust reasoning paths. In contrast, PRMs~\cite{li2022making, lightman2023let, uesato2022solving, wang2023math, wang2024multi, luo2024improve} offer a more fine-grained approach by evaluating each reasoning step, enabling more accurate and interpretable feedback.

Some recent works have explored process reward models for mathematical reasoning in pure text. PRM800k~\cite{lightman2023let} manually constructs a large-scale dataset with step-level correctness labels, which is hard to scale. MathShepherd~\cite{wang2023math} adopts Monte Carlo estimation to label reasoning steps by evaluating whether continuations from a given step lead to the correct answer, but its efficiency is relatively low. OmegaPRM~\cite{luo2024improve} introduces a Monte Carlo Tree Search (MCTS)-based framework that enables efficient automated generation of process supervision data. However, all these works focus on mathematical reasoning in pure text. In the field of multimodal mathematical reasoning, how to design an efficient framework for process supervision data generation and stably train process reward models remains a challenging problem.

To address these challenges, we propose \textbf{MM-PRM}, a powerful process reward model that effectively handles both in-domain and out-of-domain problems. Specifically, we first collect a large-scale, high-quality reasoning dataset and refine its format to train a policy model called \textbf{MM-Policy} with strong reasoning abilities and well-structured output. Inspired by OmegaPRM, we then establish an efficient multimodal process annotation framework. We first collect 10k high-quality K-12-level multimodal mathematical reasoning samples, named \textbf{MM-K12}. Starting with only 10k math problems as seed data, our framework automatically generates over 700k step-level annotations without any human supervision. Finally, we train MM-PRM with these data and evaluate its performance on multiple benchmarks using the BoN approach.

MM-PRM demonstrates strong performance and generalization across multiple benchmarks. Although we only use process data generated from K-12-level math problems for training, MM-PRM achieves remarkable results with MM-Policy on benchmarks such as MathVista that improves from 62.93\% to 67.60\%. In addition, even though MM-PRM is trained on data produced by MM-Policy, it still delivers competitive results on other models like InternVL2.5-8B and InternVL2.5-78B. For instance, on the self-collected MM-K12 test set, InternVL2.5-8B improves from 27.01\% to 37.80\%, and on MathVerse, InternVL2.5-78B improves from 50.18\% to 54.47\%. These results highlight MM-PRM’s ability to generalize both across datasets and across model size, despite being trained on limited and fixed data sources.

Finally, we provide a detailed discussion on key factors in PRM training, including learning rate and the choice between soft labels and hard labels. We find that the following factors are essential for stable PRM training: (1) using a small learning rate to ensure stable optimization; (2) adopting soft labels instead of hard thresholding to reduce noise and preserve step-wise uncertainty.

In summary, our contributions are as follows:

\begin{itemize}[leftmargin=10pt]
    \item We collect and release \textbf{MM-K12}, a curated multimodal math dataset containing 10,000 seed problems and 500 test problems, all verified to contain unique, checkable answers.
    \item We build a policy model named \textbf{MM-Policy} using a large corpus of mathematical reasoning data, and develop a fully automated, MCTS-based pipeline for scalable process supervision generation. Leveraging MM-Policy and only 10k seed problems from MM-K12, our framework produces over 700k step-level annotations without human supervision. Based on this, we train \textbf{MM-PRM}, which demonstrates strong performance across multiple benchmarks—for example, boosting accuracy on the MM-K12 test set from 33.92\% to 42.80\%, on MathVista from 62.93\% to 67.60\%, and on OlympiadBench from 15.41\% to 24.00\%.
    \item We provide a detailed discussion on the settings of key parameters such as the type of process labels and the learning rate. We summarize the key factors for stable PRM training, which can help the community train PRM models more effectively.
\end{itemize}

\section{Related work}

\subsection{Mathematical reasoning and reward modeling in LLMs}

Mathematical reasoning has become a focal point in evaluating the deep logical abilities of LLMs.~\cite{cobbe2021training, hendrycks2021measuring}. Unlike general language tasks, mathematical problems demand precise multi-step reasoning and logical coherence, motivating approaches like Chain-of-Thought (CoT) prompting\cite{wei2022chain} and Self-Consistency sampling~\cite{wang2022self}. Supervised fine-tuning (SFT) with structured solution datasets further reinforces reasoning performance, as demonstrated by models like Qwen~\cite{bai2023qwen}, InternLM~\cite{2023internlm}, and Gemini~\cite{team2023gemini}. More recently, reinforcement learning (RL)-based optimization, exemplified by OpenAI’s o1~\cite{jaech2024openai} and DeepSeek R1~\cite{guo2025deepseek}, has emerged as a powerful strategy that surpassing SFT-only baselines.

However, despite these advances, current LLMs still frequently produce logically inconsistent reasoning steps or false-positive solutions. To address this issue, reward modeling techniques have emerged. Traditional Outcome Reward Models (ORMs)\cite{cobbe2021training, yu2023ovm}, which assign rewards based solely on the final answer, fail to detect flawed intermediate reasoning. Process Reward Models (PRMs)\cite{lightman2023let, wang2023math, wang2024multi, luo2024improve} overcome this limitation by explicitly providing step-level supervision, significantly improving logical coherence. Representative works such as PRM800K~\cite{lightman2023let}, MathShepherd~\cite{wang2023math}, and MiPS~\cite{wang2024multi} have demonstrated the effectiveness of PRMs in enhancing mathematical reasoning. During the preparation of this manuscript, we became aware of a concurrent effort, URSA~\cite{luo2025ursa}, which also explores PRM training. While both approaches share some methodological similarities in PRM construction, our work places greater emphasis on building a scalable PRM pipeline and analyzing training dynamics, whereas URSA focuses more on integrating PRMs into RL.

Overall, PRMs have become an essential approach for improving the robustness and logical consistency of LLM reasoning through fine-grained, step-level evaluation.

\subsection{Process supervision data construction}

Training PRMs requires large-scale, high-quality step-level annotations that indicate the correctness of each intermediate reasoning step. However, most publicly available datasets focus solely on final answers, making it challenging to obtain supervision signals for reasoning quality. Existing process supervision construction methods can be broadly categorized into three classes:

\textbf{Manual Annotation.}
The earliest efforts, such as PRM800K~\cite{lightman2023let}, relied on human experts to label each reasoning step. While this yields high-quality annotations, the approach is costly, time-consuming, and difficult to scale---particularly for domains like mathematics where expert knowledge is required.

\textbf{Monte Carlo Estimation.}
Works like MathShepherd~\cite{wang2023math} and MiPS~\cite{wang2024multi} automate supervision by sampling multiple rollouts from each reasoning prefix and estimating step correctness based on the proportion of successful completions. If continuations from a given step often lead to correct answers, that step is considered reliable. This method is simple but suffers from label instability due to high variance in sampled rollouts---especially on long or complex reasoning chains.

\textbf{Monte Carlo Tree Search (MCTS).}
To address these limitations, OmegaPRM~\cite{luo2024improve} introduces a structured, search-based alternative. Using a divide-and-conquer MCTS algorithm, it efficiently identifies the first error in a reasoning chain and constructs a tree of reasoning paths with dynamically updated statistics. This approach improves annotation stability, better handles deep reasoning tasks, and generates large volumes of fine-grained supervision from limited seed data without human input.

Each method reflects a trade-off between cost, scalability, and label fidelity. Among them, MCTS-based approaches have recently emerged as the most effective strategy for generating robust and scalable process supervision, especially for multi-step mathematical reasoning tasks.

\section{Method}

\subsection{Overview}

\begin{figure*}[t]
    \centering
    \includegraphics[width=1.0\linewidth]{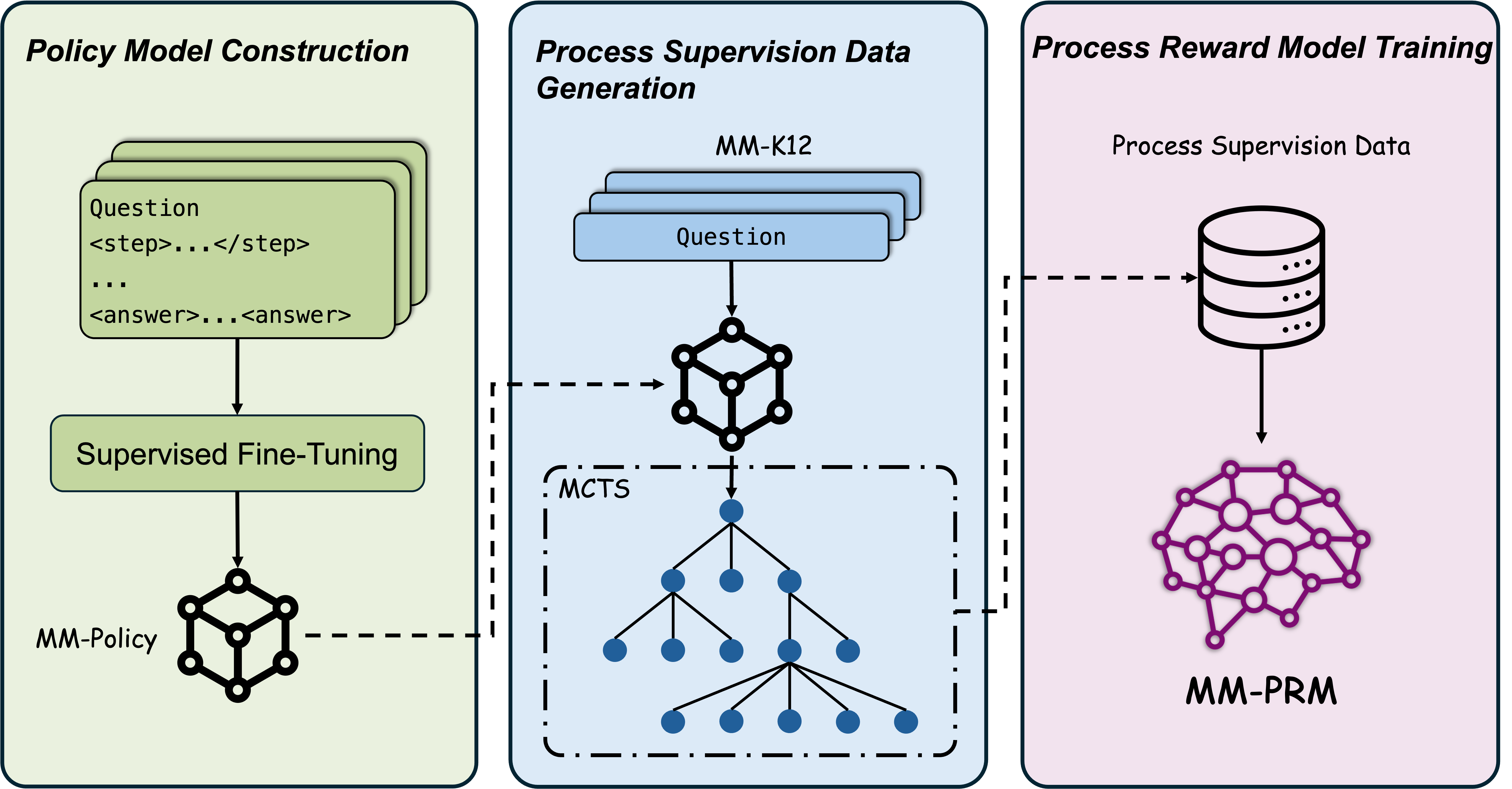}
    \caption{Overview of our automated three-stage framework for training MM-PRM: supervised policy model construction, MCTS-based process supervision data generation, and step-level reward model training. }
    \label{fig:overview}
\end{figure*}

We present a scalable and fully automated framework for training PRMs tailored to multimodal mathematical reasoning. Our approach addresses two critical bottlenecks in prior work: \textbf{the lack of step-level supervision} and the \textbf{inefficiency of data generation in complex reasoning tasks}. To overcome these challenges, we introduce a structured three-stage pipeline that enables fine-grained reward modeling without any human annotation.
As shown in Figure~\ref{fig:overview}, the framework comprises three interconnected stages as follows:

(1) \textbf{Policy Model Construction}, where a multimodal policy model is trained to generate high-quality reasoning traces following the CoT paradigm.

(2) \textbf{Process Supervision Data Generation}, where we use a MCTS-based engine, OmegaPRM~\cite{luo2024improve}, to efficiently identify reasoning flaws and produce step-level reward labels at scale.

(3) \textbf{Process Reward Model Training}, where a PRM is trained to evaluate each reasoning step and provide dense feedback.

This end-to-end design ensures that process supervision can be generated, modeled, and applied in a fully closed loop. It significantly improves reasoning quality and robustness, especially in tasks requiring long logical chains.  In Sections~\ref{sec: Policy model construction}--\ref{sec: Process supervision data generation} below, we elaborate on these three stages.

\subsection{Policy model construction}\label{sec: Policy model construction}

The policy model serves as the foundation of our framework, responsible for generating candidate reasoning trajectories given multimodal math problems. These trajectories are later evaluated and labeled to form step-level supervision for training the PRM. Ensuring that the policy model produces logically coherent and structurally complete outputs is thus essential for the effectiveness of the entire system.

To train the policy model, we curated a large-scale, high-quality dataset of mathematical problems spanning a wide range of topics and difficulty levels. Our dataset integrates samples from over a dozen public math datasets, including R-CoT~\cite{deng2024r}, MAVIS~\cite{zhang2024mavis}, MathV360K~\cite{shi2024math}, NuminaMath~\cite{li2024numinamath}, and DART-Math~\cite{tong2024dart}, with problems ranging from elementary school arithmetic to advanced geometry and statistics. The full list of data sources and sample counts used for training is provided in Appendix~\ref{apx: data_source}.

Once collected, all data undergo rigorous cleaning and format standardization. Visual and textual content are paired explicitly, and reasoning traces are reformatted to follow a structured CoT schema, with each logical step clearly marked using structured tags such as \texttt{<step>}\texttt{</step>}, and the final conclusion labeled with \texttt{<answer>}\texttt{</answer>}. To enhance quality and clarity, we leveraged a strong instruction-tuned language model (i.e., Qwen2.5-72B-Instruct~\cite{bai2023qwen}) to parse original solutions and restructure them into coherent, modular steps. This structured representation not only enhances model learnability but also lays the foundation for generating step-level reward labels in the next stage. The prompt used for solution restructuring is provided in Appendix~\ref{apx:prompt}.

With this cleaned and annotated corpus of over 5 million examples, we fine-tuned a strong open-source multimodal model, InternVL2.5-8B~\cite{chen2024internvl}, using supervised learning. This ensures that the model learns to produce logically sound and well-structured outputs that conform to the CoT reasoning pattern.

The resulting model not only delivers high-quality reasoning trajectories for downstream data annotation but also supports inference-time use cases such as BoN selection with reward-based reranking. Its strong generative ability and logical consistency form the cornerstone of our scalable process supervision framework.

\subsection{Process supervision data generation}\label{sec: Process supervision data generation}

To enable fine-grained supervision for step-level reasoning, we adopt an automated process annotation pipeline based on the OmegaPRM~\cite{luo2024improve} framework. OmegaPRM introduces a MCTS-based mechanism for efficiently identifying and labeling intermediate reasoning steps with confidence estimates. Although originally developed for textual mathematical reasoning, we adapt and extend this framework to handle multimodal inputs.

Our process begins with collecting a curated dataset of 10,000 multimodal math problems from real-world named \textbf{MM-K12}—consisting of 5,000 fill-in-the-blank and 5,000 multiple-choice questions. These problems span a range of curriculum topics from elementary to high school and serve as seed instances for process supervision generation. All examples in MM-K12 are carefully filtered to ensure that each question includes meaningful visual input and has a unique, verifiable answer, making them well-suited for structured reasoning and reward modeling. In addition, MM-K12 also provides an independent test set of 500 problems, constructed under the same criteria, which we use to evaluate in-distribution performance later. For each problem, the policy model produces multiple candidate solutions following the CoT paradigm, and these reasoning paths form the raw material for subsequent reward annotation.

To evaluate the correctness of each intermediate step, we follow OmegaPRM's hierarchical rollout and search protocol. Specifically, we generate multiple completions (rollouts) from partial prefixes and estimate the correctness of a given step based on whether its downstream completions reach the correct final answer. By applying binary search, the algorithm efficiently pinpoints the earliest step at which the reasoning begins to deviate. These supervision signals are then aggregated within a structured state-action tree, which records the Monte Carlo (MC) estimates and other statistics at each reasoning state. In our implementation, we maintain the full multimodal context---including both textual and visual components---throughout the tree construction and search process.

Importantly, our adaptation retains the efficiency of OmegaPRM's divide-and-conquer search while enabling reward supervision for reasoning steps that are conditioned on complex visual stimuli. Through this pipeline, we generate over 700,000 step-level annotations from only 10k seed questions, without requiring manual labeling. The resulting dataset provides dense, high-quality process supervision aligned with real-world multimodal reasoning.

\subsection{Process reward model training}\label{sec: Process reward model training}

With large-scale step-level supervision in place, we proceed to train a PRM that can assess the quality of reasoning steps given a multimodal context. The PRM is designed to serve as a fine-grained critic, assigning a reward score to each intermediate step conditioned on its preceding reasoning context that enables both test-time scaling and potential RL applications.

\subsubsection{Labeling strategy}\label{sec: labeling strategy}

A central design decision in PRM training lies in how to formulate supervision signals from the MC estimations~\cite{luo2024improve, zhang2025lessons}. Rather than adopting hard binary label (e.g., $\hat{y} = 1[\mathrm{MC}(s) > \tau]$), we use soft label, directly taking the MC scores as continuous supervision targets.

This choice is motivated by the observation that the MC score reflects more than the correctness of an intermediate step. It also encodes factors such as problem difficulty, step criticality, and distributional uncertainty in the policy model’s rollouts. For instance, a reasoning step within a highly ambiguous or visually complex problem may yield lower MC scores even if the logic is fundamentally sound. In such cases, hard-thresholding may misrepresent the step’s quality, introducing noise into training. By contrast, soft labels preserve the probabilistic nuance and enable smoother learning dynamics, more details will be discussed in \sectionautorefname~\ref{sec: labeling strategy discussion}

Formally, for each reasoning step $x_t$ in a path $x = [x_1, x_2, \dots, x_T]$, we assign a supervision target $\hat{y}_t = \mathrm{MC}(x_{<t}) \in [0, 1]$, where $\mathrm{MC}(x_{<t})$ denotes the estimated probability that a correct final answer can be reached from this partial path.

\subsubsection{Model design and training objective}

To model the prediction task, we treat the PRM as a classifier operating on each reasoning step. Given a multimodal input $q$ and a generated reasoning trace $[x_1, x_2, \dots, x_T]$, we interleave a special marker token, denoted $\sigma$, after each step, producing an input sequence of the form:
\[
[q, x_1, \sigma, x_2, \sigma, \dots, x_T, \sigma].
\]
In our implementation, $\sigma$ is instantiated as the token \texttt{<prm>}. At each occurrence of $\sigma$, the model is tasked with producing a scalar confidence score indicating the likelihood that the immediately preceding step is logically correct. Formally, let $z^{(i)}_{\texttt{Yes}}$ and $z^{(i)}_{\texttt{No}}$ denote the unnormalized logits for binary labels “Yes” (correct) and “No” (incorrect) at the $i$-th occurrence of $\sigma$. The model's predicted probability is computed via softmax:
\[
p^{(i)} = \frac{\exp(z^{(i)}_{\texttt{Yes}})}{\exp(z^{(i)}_{\texttt{Yes}}) + \exp(z^{(i)}_{\texttt{No}})}.
\]
The training objective is to minimize the \textbf{cross-entropy loss} between the predicted scores $p^{(i)}$ and the soft labels $\hat{y}^{(i)}$, across all scoring points:
\[
\mathcal{L}_{\text{PRM}} = - \sum_{i=1}^{T} \left[ \hat{y}^{(i)} \cdot \log p^{(i)} + (1 - \hat{y}^{(i)}) \cdot \log(1 - p^{(i)}) \right].
\]
This formulation guides the model to make fine-grained assessments of reasoning steps, assigning higher confidence to those with stronger evidence of correctness.

\section{Experiments}

\subsection{Experiments setup}\label{sec: experiment setup}

To validate the effectiveness of our proposed process reward modeling framework, we conduct a series of experiments, carefully configured to ensure fair, scalable, and reproducible results.

\textbf{Policy model construction.}
Our policy model(i.e., MM-Policy) is initialized from the multimodal backbone InternVL 2.5-8B and fine-tuned using approximately 4 million cleaned, structured math problems. The model is trained for 1 epoch with a batch size of 128 and a learning rate of 4e-5, updating only the language module while keeping the vision encoder frozen.

\textbf{Process supervision data generation.}
We adapt the OmegaPRM\cite{luo2024improve} pipeline for multimodal reasoning and apply it to MM-K12 (10k samples). Using MCTS-based structured rollouts, we generate approximately 747,000 step-level annotations. The sampling parameters are tuned for balance between diversity and efficiency: $temperature = 1.0$, $top_k = 50$, $top_p = 0.9$, exploration coefficient $c_{\text{puct}} = 0.125$, and up to 200 search steps or 1,000 total rollouts per problem.

\textbf{Process reward model training.}
We initialize the PRM from the fine-tuned policy model and train it for 1 epochs with a batch size of 512 and a learning rate of 4e-6.

\subsection{Evaluation strategies and benchmarks}

To assess the effectiveness of the proposed MM-PRM in improving reasoning quality, we adopt the BoN evaluation protocol. For each test problem, the policy model generates $N = 16$ candidate reasoning paths independently. The PRM then scores each path step-by-step, producing a sequence of floating-point values representing the predicted quality of each intermediate step, the path with the highest score is selected as the final answer.

Since PRM outputs a vector of step-wise confidence scores for each candidate path, a crucial component of our evaluation is the \textit{aggregation function}~\cite{wang2024multi} used to compress this vector into a scalar. We explore a diverse set of aggregation functions, including \texttt{Min}, \texttt{Average}, \texttt{Max}, \texttt{SumLogPr} (i.e., sum of log-probabilities), \texttt{SumLogOdds} (i.e., sum of log-odds), and \texttt{MeanOdds} (i.e., mean odds), each function captures different aspects of path quality. Additionally, a \texttt{Random} baseline is used for comparison, where the final answer is randomly sampled from the same set of 16 candidates. Formal definitions of all aggregation functions are provided in Appendix~\ref{apx:aggregation}.

We evaluate performance using answer accuracy, defined as the proportion of final selected answers that match the ground truth. This metric directly reflects MM-PRM's utility in guiding the selection of correct reasoning paths. To comprehensively evaluate our model's performance and generalization, we conduct experiments on a range of multimodal math benchmarks, including MM-K12 (test set), OlympiadBench (OE\_MM\_maths\_en\_COMP)~\cite{he2024olympiadbench}, MathVista(testmini)~\cite{lu2023mathvista}, MathVerse(testmini)~\cite{zhang2024mathverse}, and MathVision(test)~\cite{wang2024measuring}. The MM-K12 test set serves as an in-distribution evaluation. For out-of-distribution assessment, we use the \texttt{OE\_MM\_maths\_en\_COMP} split of OlympiadBench, which contains open-ended multimodal questions from international math competitions, closely related in format to MM-K12 but significantly harder. To further test generalization, we include MathVista, which covers a wide range of visual mathematical tasks; MathVerse, which emphasizes understanding structured visual content; and MathVision, which targets abstract visual reasoning. These benchmarks provide a diverse and rigorous setting to measure both performance and generalization of our process reward modeling framework.

\subsection{Quantitative results}

\begin{table}[ht]
    \centering
    \caption{Performance improvements across various benchmarks when applying the MM-PRM to different models. }
    \label{tab:generalization}
    \begin{tabular}{lcccccc}
        \toprule
        \textbf{Model} & \textbf{MM-K12} & \textbf{OlympiadBench} & \textbf{MathVista} & \textbf{MathVerse} & \textbf{MathVision} \\
        \midrule
        MM-Policy & 33.92 & 15.41 & 62.93 & 42.99 & 21.74\\
        +\textit{MM-PRM} & 42.80 & 24.00 & 67.60 & 46.27 & 27.11\\
         & \cellcolor{RoyalBlue!5}{\textbf{+8.88}} & \cellcolor{RoyalBlue!5}{\textbf{+8.59}} & \cellcolor{RoyalBlue!5}{\textbf{+4.67}} & \cellcolor{RoyalBlue!5}{\textbf{+3.28}} & \cellcolor{RoyalBlue!5}{\textbf{+5.37}} \\
        \midrule
        InternVL2.5-8B & 27.01 & 5.23 & 56.43 & 36.26 & 10.04 \\
        +\textit{MM-PRM} & 37.80 & 15.33 & 63.50 & 42.56 & 19.41 \\
        & \cellcolor{RoyalBlue!5}{\textbf{+10.79}} & \cellcolor{RoyalBlue!5}{\textbf{+10.10}} & \cellcolor{RoyalBlue!5}{\textbf{+7.07}} & \cellcolor{RoyalBlue!5}{\textbf{+6.30}} & \cellcolor{RoyalBlue!5}{\textbf{+9.37}} \\
        \midrule
        InternVL2.5-26B & 28.01 & 14.46 & 60.02 & 37.83 & 20.76 \\
        +\textit{MM-PRM} & 38.00 & 24.67 & 64.50 & 44.19 & 25.63 \\
        & \cellcolor{RoyalBlue!5}{\textbf{+9.99}} & \cellcolor{RoyalBlue!5}{\textbf{+10.21}} & \cellcolor{RoyalBlue!5}{\textbf{+4.25}} & \cellcolor{RoyalBlue!5}{\textbf{+6.36}} & \cellcolor{RoyalBlue!5}{\textbf{+4.87}} \\
        \midrule
        InternVL2.5-38B & 40.34 & 29.57 & 68.32 & 47.94 & 29.70 \\
        +\textit{MM-PRM} & 52.40 & 32.67 & 71.10 & 52.61 & 32.99 \\
        & \cellcolor{RoyalBlue!5}{\textbf{+12.06}} & \cellcolor{RoyalBlue!5}{\textbf{+3.10}} & \cellcolor{RoyalBlue!5}{\textbf{+2.78}} & \cellcolor{RoyalBlue!5}{\textbf{+4.67}} & \cellcolor{RoyalBlue!5}{\textbf{+3.29}} \\
        \midrule
        InternVL2.5-78B & 42.24 & 30.98 & 69.48 & 50.18 & 31.50 \\
        +\textit{MM-PRM} & 48.80 & 34.67 & 73.20 & 54.47 & 33.26 \\
        & \cellcolor{RoyalBlue!5}{\textbf{+6.56}} & \cellcolor{RoyalBlue!5}{\textbf{+3.69}} & \cellcolor{RoyalBlue!5}{\textbf{+3.72}} & \cellcolor{RoyalBlue!5}{\textbf{+4.29}} & \cellcolor{RoyalBlue!5}{\textbf{+1.76}} \\
        \bottomrule
    \end{tabular}
\end{table}

We evaluate the effectiveness of MM-PRM by applying it to a range of policy models and testing its impact across multiple multimodal math benchmarks. 

Across all models, MM-PRM yields substantial performance improvements. For example, when applied to MM-Policy on the MM-K12 test set, accuracy improves from 33.92\% to 42.80\%, and similar gains are observed with InternVL2.5-8B, where performance rises from 27.01\% to 37.80\%. These results confirm that MM-PRM is highly effective at identifying high-quality reasoning paths.

Beyond in-domain settings, we observe that MM-PRM also generalizes well to larger models and more challenging datasets. As shown in Table~\ref{tab:generalization}, applying MM-PRM to InternVL2.5-78B improves accuracy on OlympiadBench from 30.98\% to 34.67\%, and on MathVerse from 50.18\% to 54.47\%. Despite being trained only on the small MM-K12 seed dataset and with a fixed policy model(i.e., MM-Policy), MM-PRM consistently enhances reasoning accuracy across diverse benchmarks and models. This demonstrates the potential of scalable step-level reward modeling to improve mathematical reasoning in a model-agnostic and data-efficient manner. Detailed evaluation results across all aggregation functions are provided in Appendix~\ref{apx:full_results}.

\subsection{Qualitative results}

\begin{figure*}[t]
    \centering
    \includegraphics[width=1.0\linewidth]{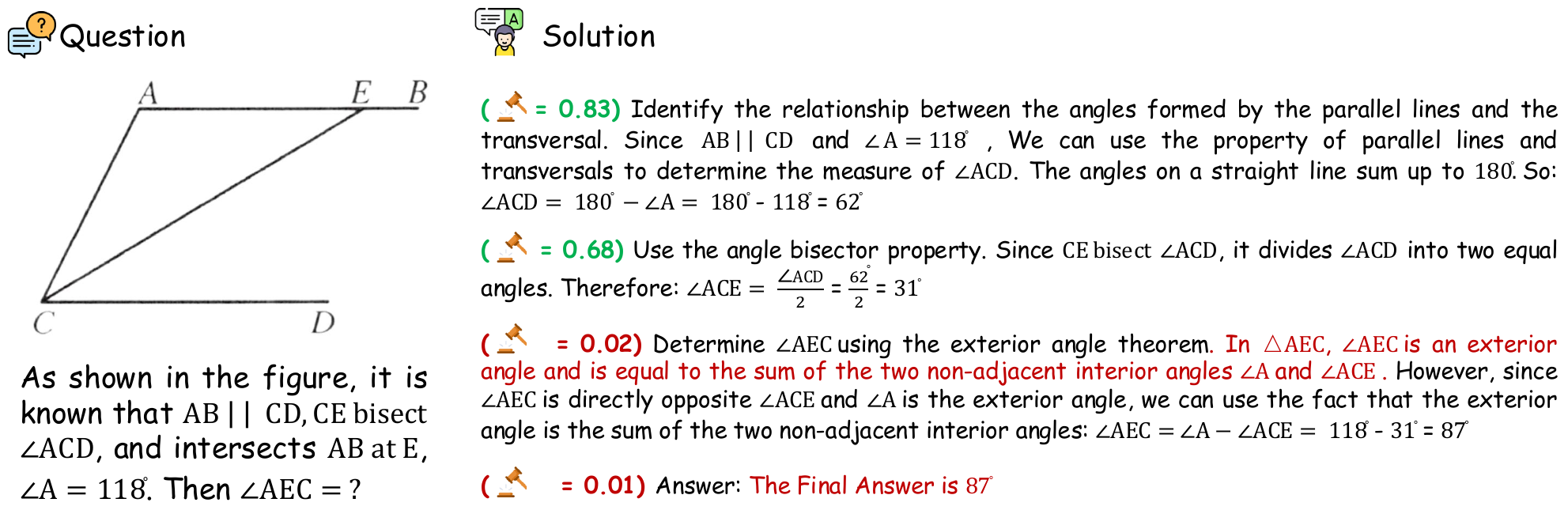}
    \caption{Qualitative example of MM-PRM accurately identifying error steps in multimodal reasoning process. }
    \label{fig:qualitative_results}
\end{figure*}

To further illustrate the behavior of MM-PRM, we present a qualitative example involving a geometry problem with parallel lines and an angle bisector. As shown in Figure~\ref{fig:qualitative_results}, the policy model generates a four-step reasoning path, which ultimately leads to an incorrect final answer.

The PRM assigns high scores to the first two steps (0.83 and 0.68), indicating that they are logically sound. In contrast, Step-3 receives a very low score (0.02), signaling that the model has identified a significant reasoning error at this point. This flawed step leads directly to an incorrect conclusion in Step-4.

This example demonstrates that MM-PRM is capable of detecting localized logical errors within a reasoning chain, such fine-grained judgment is crucial in selecting high-quality reasoning trajectories and filtering out those with subtle but critical flaws.

\section{Discussion}

\begin{figure*}[ht]
    \centering
    \includegraphics[width=1.0\linewidth]{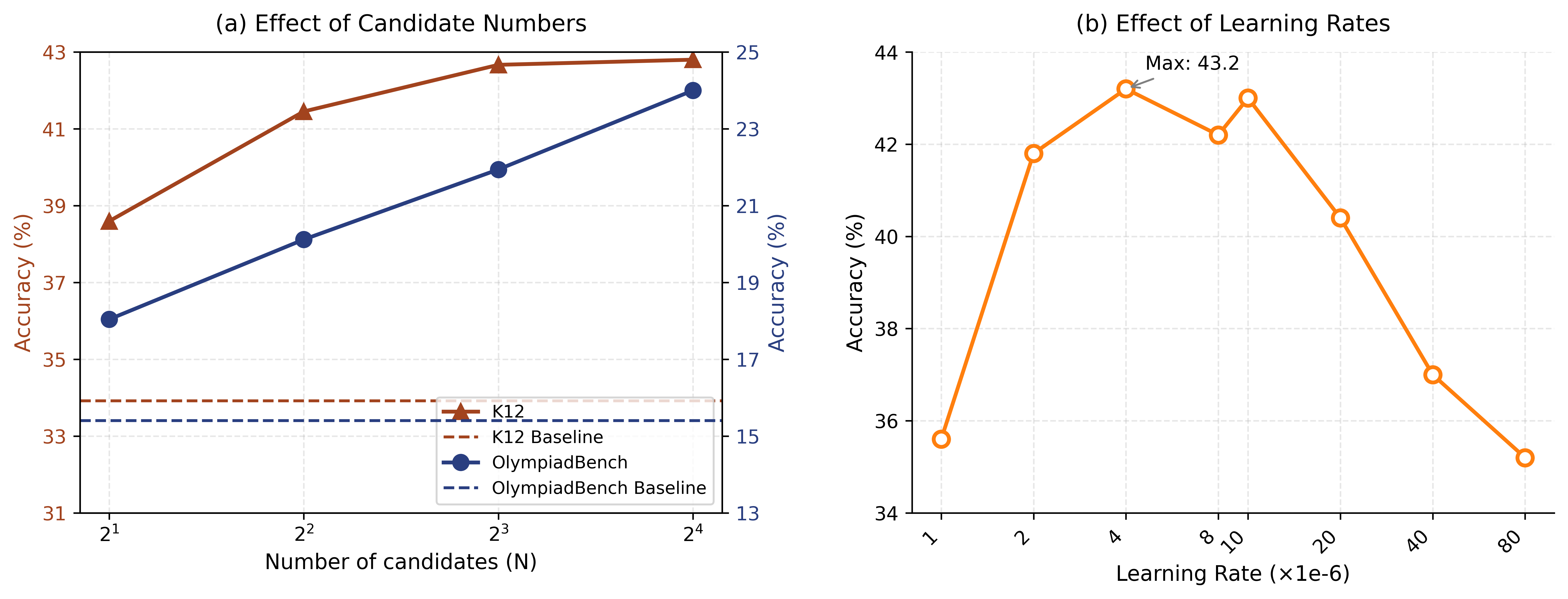}
    \caption{(a) Effect of the number of candidate reasoning paths on answer accuracy under Best-of-N inference. Increasing the number of candidates allows the PRM to select higher-quality reasoning trajectories. (b) Effect of learning rate on PRM performance. Small learning rates yield better accuracy, with performance peaking at 4e-6}
    \label{fig:combined_plot}
\end{figure*}

\subsection{Candidate path's impact on PRM performance}


Since the PRM operates purely as a selector in the BoN framework, its performance is inherently bounded by the diversity and quality of candidate reasoning paths produced by the policy model. In other words, PRM cannot improve a flawed generation in BoN---it can only choose among the available options. Therefore, the number of reasoning paths generated per problem directly affects its potential to identify a correct and coherent solution.

To study this effect, we vary the number of generated reasoning paths $N \in \{2, 2^2, 2^3, 2^4\}$ and measure the corresponding answer accuracy under the aggregation strategy \texttt{MeanOdds}. As shown in Figure~\ref{fig:combined_plot}(a), increasing $N$ consistently improves MM-PRM's performance across both test sets.
On the MM-K12 test set, accuracy improves from 38.6\% at $N = 2$ to 42.8\% at $N = 16$, with gains tapering off beyond $N = 8$.
In contrast, on OlympiadBench, accuracy increases more steadily—from 18.4\% to 24.0\%---as $N$ grows. This suggests that for harder, more diverse tasks, having a larger pool of reasoning paths is critical for PRM to identify valid solutions.

\subsection{Learning rate}

As noted in~\cite{lightman2023let}, finetuning a PRM shifts the language models' objective from generation to discrimination, making learning rate a critical factor. Smaller learning rates are often preferred to maintain stability and preserve pretrained knowledge.

We evaluate MM-PRM trained under different learning rates on the MM-K12 test set using \texttt{MeanOdds} aggregator. As shown in Figure~\ref{fig:combined_plot}(b), performance peaks at 4e-6---about one-tenth the learning rate typically used in supervised fine-tuning---then drops sharply at higher values. This confirms that a moderate, conservative learning rate leads to better training, while overly large values degrade accuracy.

\subsection{Soft label vs. Hard label}\label{sec: labeling strategy discussion}

\begin{table}[ht]
    \centering
    \caption{The performance comparisons of soft label vs. hard label. }
    \label{tab:soft vs hard}
    \begin{tabular}{lcccccccc}
        \toprule
        \textbf{Labeling Strategy} & \texttt{Min} & \texttt{Average} & \texttt{Max} & \texttt{SumLogPr} & \texttt{SumLogOdds} & \texttt{MeanOdds} \\
        \midrule
        Soft Label & 37.4 & 43.0 & 43.4 & 42.0 & 43.2 & 42.8 \\
        Hard Label & 36.8 & 34.4 & 35.6 & 36.0 & 33.8 & 37.0 \\
        \bottomrule
    \end{tabular}
\end{table}

As discussed in \sectionautorefname~\ref{sec: labeling strategy}, we adopt soft label---i.e., real-valued MC scores---as supervision for step-level reward modeling. Unlike hard label, soft label retain uncertainty and allow the model to learn more nuanced representations of reasoning quality.

To assess this design choice, we compare soft label with hard label thresholding, where steps with $MC > 0$ are treated as correct, and others as incorrect, following the protocol in~\cite{luo2024improve, wang2024openr}.

As shown in Table~\ref{tab:soft vs hard}, soft-label training consistently outperforms hard-label training across all aggregation strategies. For instance, under the \texttt{Average} aggregator, soft labels yield 43\% accuracy on MM-K12 test set, compared to 34.4\% with hard labels. Similar improvements are observed with \texttt{SumLogOdds} (43.2\% vs. 33.8\%) and \texttt{MeanOdds} (42.8\% vs. 37.0\%).



\section{Conclusion}

In this work, we introduce MM-PRM, a process reward model built upon a scalable framework for multimodal mathematical process reward modeling that enables step-level supervision without human annotation. By leveraging a multimodal policy model and an MCTS-based data generation pipeline, we construct over 700k process-level labels from only 10k K-12 math problems in our MM-K12 dataset. Our trained PRM significantly improves reasoning accuracy in BoN inference across both standard and challenging benchmarks, and demonstrates strong generalization to new datasets and models. Extensive analysis further confirms the importance of soft labeling, conservative learning rates, and sufficient path diversity. MM-PRM highlights the value of process supervision for enhancing multimodal mathematical problem-solving.



{
\small
\bibliographystyle{unsrt}
\bibliography{neurips_2025}
}

\newpage
\appendix

\section*{Appendix}

\section{Policy Model training data}\label{apx: data_source}

Table~\ref{tab:data_source} summarizes the datasets and sample counts used to train the policy model.

\begin{table}[H]
    \centering
    \caption{Data sources and sample counts used for policy model training.}
    \label{tab:data_source}
    \scalebox{1.0}{
    \begin{tabular}{l r}
        \toprule
        \textbf{Dataset} & \textbf{Sample Count} \\
        \midrule
        R-CoT~\cite{deng2024r} & 201,781 \\
        MAVIS~\cite{zhang2024mavis} & 29,551 \\
        multimath-300k~\cite{peng2024multimath} & 224,747 \\
        Multimodal ArXiv~\cite{li2024multimodal} & 99,772 \\
        MathV360K~\cite{shi2024math} & 338,721 \\
        Math-PUMA~\cite{zhuang2025math} & 1,111,939 \\
        Multi-modal-Self-instruct~\cite{zhang2024multimodal} & 64,807 \\
        MMPR~\cite{wang2024mpo} & 1,055,796 \\
        ShareGPT-4o~\cite{cui2025comprehensive} & 57,289 \\
        table-vqa~\cite{AgDeTQA} & 80,054 \\
        MathQA~\cite{amini2019mathqa} & 27,923 \\
        GSM8K~\cite{cobbe2021training} & 7,413 \\
        DART-Math~\cite{tong2024dart} & 574,305 \\
        math-gpt-4o-200k~\cite{pawankrd2024math} & 196,032 \\
        NuminaMath-CoT~\cite{li2024numinamath} & 807,883 \\
        MathInstruct~\cite{yue2023mammoth} & 257,755 \\
        \midrule
        \textbf{Total} & \textbf{5,135,768} \\
        \bottomrule
    \end{tabular}
    }
\end{table}

\section{Prompt template for data cleaning} \label{apx:prompt}
The prompt template used for data cleaning is shown below. Before generating logical steps and the final answer, the language model is first instructed to explicitly identify the key knowledge points involved in the problem, along with brief explanations. This two-stage prompting strategy enhances the model’s understanding of the question prior to solution restructuring.

\begin{tcolorbox}[breakable]
\begin{verbatim}
Using the information provided, identify and summarize the key 
knowledge points required to solve the problem and rewrite the
original answer with a detailed reasoning process based on the
input answer.
Provide a clear explanation of each knowledge point by stating
its name followed by a colon (:), and then presenting the detailed 
Explanation on the same line.
When explaining the knowledge point, DO NOT reference or describe 
the original question, answer, or answer details. Focus solely on
explaining the knowledge points.
Besides, rewrite the original answer to include detailed reasoning
and answer. Do not change the final answer and always refer to the
input answer.

Output Format:
In order to answer this question, we first need to have the 
following prior knowledge:
{{Substitute with name of knowledge point 1}}: {{Substitute with
Explanation of knowledge point 1}},
{{Substitute with name of knowledge point 2}}: {{Substitute with
Explanation of knowledge point 2}},
...
We answer this based on prior knowledge as follows:
Solution: Refined answer with detailed reasoning. Use Step 1, Step 2 
to divide the steps. Remember do not change the final answer and
always refer to the input answer.
Answer: The Final Answer is {{Substitute with final answer}}. 

Input Information:

Question: {question} 
--------------------------------
Answer: {answer}
--------------------------------
\end{verbatim}
\end{tcolorbox}

\section{Aggregation function definitions} \label{apx:aggregation}

In the BoN inference setup, each reasoning path is assigned a vector of step-level scores predicted by the PRM. To compare and rank these paths, we apply an aggregation function to compress each score vector into a single scalar value, following the design considerations discussed in~\cite{wang2024multi}. The following aggregation strategies are used in our experiments:

\begin{itemize}
    \item \texttt{Min}: Computes the minimum of all step scores:
    \[
    \text{score}(r_j) = \min \{p_1^{(j)}, p_2^{(j)}, \dots, p_{T_j}^{(j)}\}
    \]

    \item \texttt{Max}: Computes the maximum of all step scores:
    \[
    \text{score}(r_j) = \max \{p_1^{(j)}, p_2^{(j)}, \dots, p_{T_j}^{(j)}\}
    \]

    \item \texttt{Average}: Computes the arithmetic mean of all step scores:
    \[
    \text{score}(r_j) = \frac{1}{T_j} \sum_{i=1}^{T_j} p_i^{(j)}
    \]

    \item \texttt{SumLogPr}: Computes the sum of log-probabilities of all step scores:
    \[
    \text{score}(r_j) = \sum_{i=1}^{T_j} \log p_i^{(j)} = \log \prod_{i=1}^{T_j} p_i^{(j)}
    \]

    \item \texttt{SumLogOdds}: Computes the sum of log-odds of all step scores:
    \[
    \text{score}(r_j) = \sum_{i=1}^{T_j} \log \frac{p_i^{(j)}}{1 - p_i^{(j)}}
    \]

    \item \texttt{MeanOdds}: Computes the mean of odds-transformed step scores:
    \[
    \text{score}(r_j) = \frac{1}{T_j} \sum_{i=1}^{T_j} \frac{p_i^{(j)}}{1 - p_i^{(j)}}
    \]
\end{itemize}

Here, \( r_j \) denotes the \( j \)-th reasoning path consisting of \( T_j \) steps, and \( p_i^{(j)} \in (0, 1) \) is the predicted correctness probability of the \( i \)-th step in that path.

\section{Full evaluation results} \label{apx:full_results}

\subsection{MM-Policy}
\begin{table}[H]
    \centering
    \caption{Performance comparison of various aggregators on various benchmarks for MM-Policy. Top performer is in \textbf{bold}. }
    \label{tab:full_results_mm-policy}
    \setlength{\tabcolsep}{5.3pt}
    \begin{tabular}{lcccccccc}
        \toprule
        \textbf{Benchmark} & \texttt{Random} & \texttt{Min} & \texttt{Average} & \texttt{Max} & \texttt{SumLogPr} & \texttt{SumLogOdds} & \texttt{MeanOdds} \\
        \midrule
    MM-K12  & \cellcolor{RoyalBlue!5}{33.9}  & 37.4  & 43.0  & \textbf{43.4 } & 42.0  & 43.2  & 42.8  \\
    OlympiadBench & \cellcolor{RoyalBlue!5}{15.4}  & 23.3  & 20.0  & \textbf{24.7 } & 22.7  & 24.0  & 24.0  \\
    MathVista & \cellcolor{RoyalBlue!5}{62.9}  & 66.0  & 67.2  & \textbf{67.7 } & 66.4  & \textbf{67.7 } & 67.6  \\
    MathVerse & \cellcolor{RoyalBlue!5}{43.0}  & 46.0  & 48.0  & 46.1  & 47.3  & \textbf{48.0 } & 46.3  \\
    MathVision & \cellcolor{RoyalBlue!5}{21.7}  & 25.8  & 26.7  & 26.9  & 25.4  & 26.3  & \textbf{27.1 } \\
        \bottomrule
    \end{tabular}
\end{table}

\subsection{InternVL-8B}
\begin{table}[H]
    \centering
    \caption{Performance comparison of various aggregators on various benchmarks for InternVL-8B. Top performer is in \textbf{bold}. }
    \label{tab:full_results_internvl8b}
    \setlength{\tabcolsep}{5.3pt}
    \begin{tabular}{lcccccccc}
        \toprule
        \textbf{Benchmark} & \texttt{Random} & \texttt{Min} & \texttt{Average} & \texttt{Max} & \texttt{SumLogPr} & \texttt{SumLogOdds} & \texttt{MeanOdds} \\
        \midrule
    MM-K12  & \cellcolor{RoyalBlue!5}{27.0}  & 34.8  & 36.0  & 35.8  & 31.2  & 34.2  & \textbf{37.8} \\
    OlympiadBench & \cellcolor{RoyalBlue!5}{5.2 } & 2.7  & 15.3  & \textbf{16.7 } & 2.0  & 1.3  & 15.3  \\
    MathVista  & \cellcolor{RoyalBlue!5}{56.4}  & 63.4  & 62.8  & 62.6  & 61.0  & 62.0  & \textbf{63.5} \\
    MathVerse & \cellcolor{RoyalBlue!5}{36.3}  & 41.8  & \textbf{42.8 } & 41.1  & 42.0  & 42.8  & 42.6  \\
    MathVision & \cellcolor{RoyalBlue!5}{10.0}  & 6.1  & 18.4  & \textbf{20.6 } & 3.8  & 4.8  & 19.4  \\
        \bottomrule
    \end{tabular}
\end{table}

\subsection{InternVL-26B}
\begin{table}[H]
    \centering
    \caption{Performance comparison of various aggregators on various benchmarks for InternVL-26B. Top performer is in \textbf{bold}. }
    \label{tab:full_results_internvl26b}
    \setlength{\tabcolsep}{5.3pt}
    \begin{tabular}{lcccccccc}
        \toprule
        \textbf{Benchmark} & \texttt{Random} & \texttt{Min} & \texttt{Average} & \texttt{Max} & \texttt{SumLogPr} & \texttt{SumLogOdds} & \texttt{MeanOdds} \\
        \midrule
    MM-K12  & \cellcolor{RoyalBlue!5}{28.0}  & \textbf{38.0 } & 37.2  & 35.4  & 32.0  & 33.0  & \textbf{38.0 } \\
    OlympiadBench & \cellcolor{RoyalBlue!5}{14.5}  & 16.0  & 23.3  & 22.0  & 16.7  & 16.7  & 24.7  \\
    MathVista & \cellcolor{RoyalBlue!5}{60.0}  & 63.9  & \textbf{64.6 } & 64.4  & 62.8  & 64.2  & 64.5  \\
    MathVerse & \cellcolor{RoyalBlue!5}{37.8}  & 42.0  & 44.2  & \textbf{44.3 } & 40.1  & 41.5  & 44.2  \\
    MathVision & \cellcolor{RoyalBlue!5}{20.8}  & 23.3  & 24.1  & 24.9  & 20.6  & 22.0  & \textbf{25.6 } \\
        \bottomrule
    \end{tabular}
\end{table}

\subsection{InternVL-38B}
\begin{table}[H]
    \centering
    \caption{Performance comparison of various aggregators on various benchmarks for InternVL-38B. Top performer is in \textbf{bold}. }
    \label{tab:full_results_internvl38b}
    \setlength{\tabcolsep}{5.3pt}
    \begin{tabular}{lcccccccc}
        \toprule
        \textbf{Benchmark} & \texttt{Random} & \texttt{Min} & \texttt{Average} & \texttt{Max} & \texttt{SumLogPr} & \texttt{SumLogOdds} & \texttt{MeanOdds} \\
        \midrule
    MM-K12  & \cellcolor{RoyalBlue!5}{40.3}  & 46.8  & 51.0  & 49.8  & 42.8  & 46.6  & \textbf{52.4 } \\
    OlympiadBench & \cellcolor{RoyalBlue!5}{29.6}  & 29.3  & 34.0  & \textbf{34.7 } & 28.0 & 30.0  & 32.7  \\
    MathVista & \cellcolor{RoyalBlue!5}{68.3}  & 70.5  & \textbf{71.1 } & 70.5  & 69.2  & 70.1  & \textbf{71.1 } \\ 
    MathVerse & \cellcolor{RoyalBlue!5}{47.9}  & 50.4  & 51.9  & 51.1  & 51.1  & 52.7  & \textbf{52.6 } \\
    MathVision & \cellcolor{RoyalBlue!5}{29.7}  & 29.1  & 32.6  & \textbf{33.6 } & 30.6 & 31.3  & 33.0  \\
        \bottomrule
    \end{tabular}
\end{table}

\subsection{InternVL-78B}
\begin{table}[H]
    \centering
    \caption{Performance comparison of various aggregators on various benchmarks for InternVL-78B. Top performer is in \textbf{bold}. }
    \label{tab:full_results_internvl78b}
    \setlength{\tabcolsep}{5.3pt}
    \begin{tabular}{lcccccccc}
        \toprule
        \textbf{Benchmark} & \texttt{Random} & \texttt{Min} & \texttt{Average} & \texttt{Max} & \texttt{SumLogPr} & \texttt{SumLogOdds} & \texttt{MeanOdds} \\
        \midrule
    MM-K12  & \cellcolor{RoyalBlue!5}{42.2}  & 46.2  & 47.2  & 47.2  & 45.0  & 46.8  & \textbf{48.8 } \\
    OlympiadBench & \cellcolor{RoyalBlue!5}{31.0}  & 33.3  & 35.3  & \textbf{34.7 } & 33.3  & \textbf{34.7 } & \textbf{34.7 } \\
    MathVista & \cellcolor{RoyalBlue!5}{69.5}  & 71.7  & 73.4  & 72.4  & \textbf{73.9 } & 73.6  & 73.2  \\
    MathVerse  & \cellcolor{RoyalBlue!5}{50.2}  & 53.4  & 55.0  & 54.2  & 54.3  & \textbf{55.1 } & 54.5  \\
    MathVision & \cellcolor{RoyalBlue!5}{31.5}  & 30.1  & 33.5  & \textbf{33.6 } & 31.8  & 32.4  & 33.3  \\
        \bottomrule
    \end{tabular}
\end{table}

\section{Limitations}\label{apx: limitation}

While our proposed framework demonstrates strong performance and generalization across multiple benchmarks, it also has several limitations: (1) Due to computational constraints, we conduct training only on the InternVL series with 8B parameters, without exploring larger models or architectures from other model families. This restricts our ability to fully assess how PRM training behavior scales with model size or generalizes across different backbones. (2) The seed data used for process supervision generation is limited in diversity, as it consists solely of K-12 math problems. As a result, the PRM may be less exposed to advanced mathematical domains or visual formats beyond the scope of standard educational settings. We leave broader model coverage and more diverse seed data construction as promising directions for future work.

\end{document}